\documentclass{article}

\usepackage{microtype}
\usepackage{graphicx}
\usepackage{subcaption}
\usepackage{booktabs} 
\usepackage{natbib}
\usepackage{xurl}
\usepackage{hyperref}

\usepackage[preprint]{mylatex}

\usepackage{amsmath}
\usepackage{amssymb}
\usepackage{mathtools}
\usepackage{amsthm}

\usepackage[capitalize,noabbrev]{cleveref}

\theoremstyle{plain}

\theoremstyle{definition}

\theoremstyle{remark}

\usepackage[textsize=tiny]{todonotes}

\icmltitlerunning{Full-Stack Domain Enhancement for Combustion LLMs: Construction and
Optimization}

\begin{document}

\twocolumn[
  \icmltitle{Full-Stack Domain Enhancement for Combustion LLMs: Construction and Optimization}

  \icmlsetsymbol{equal}{*}
  \icmlsetsymbol{corresponding}{†}

  \begin{icmlauthorlist}
    \icmlauthor{Quanjia Xiao}{pku}
    \icmlauthor{Weimin Ouyang}{aisi}
    \icmlauthor{Zonglin Yang}{pku}
    \icmlauthor{Tianhao Wu}{aisi}
    \icmlauthor{Qingguo Zhou}{dp}
    \icmlauthor{Runze Mao}{corresponding,pku,aisi}
    \icmlauthor{Zhi X. Chen}{corresponding,pku,aisi}
  \end{icmlauthorlist}

\icmlaffiliation{pku}{Peking University}
\icmlaffiliation{aisi}{AI for Science Institute, Beijing}
\icmlaffiliation{dp}{DP Technology}
\icmlcorrespondingauthor{Runze Mao}{maorz@pku.edu.cn}
\icmlcorrespondingauthor{Zhi X. Chen}{chenzhi@pku.edu.cn}

\icmlkeywords{Large Language Models, Domain Adaptation, Combustion Science, AI for Science}

\vskip 0.3in
]

\printAffiliationsAndNotice{\textsuperscript{†}Equal contribution corresponding authors.}

\begin{abstract}
Large language models (LLMs) in the direction of task adaptation and capability enhancement for professional fields demonstrate significant application potential.  Nevertheless, for complex physical systems such as combustion science, general-purpose LLMs often generate severe hallucinations due to insufficient domain knowledge and the inability to adhere to physical conservation laws. To address this issue, we propose the first full-stack domain-enhanced LLM workflow tailored for the field of combustion science, which integrates automated domain corpus construction, incremental pre-training, instruction fine-tuning, and verifiable reward-based reinforcement learning. This workflow ensures that the model truly internalizes physical laws rather than merely learning textual statistical patterns. We also release FlameBench, a standardized evaluation benchmark specifically designed for complex reasoning tasks in combustion science. Experimental results demonstrate that the model developed in this work significantly outperforms state-of-the-art general-purpose closed-source models and traditional retrieval-augmented generation methods on combustion science reasoning tasks. This work lays a solid technical and resource foundation for the subsequent development of domain-specific scientific research agents with reliable scientific reasoning capabilities.
\end{abstract}
\section{Introduction}
AI for Science (AI4S) has demonstrated remarkable success in advancing model construction and problem-solving for specialized fields, exemplified by advances in protein structure prediction~\citep{jumper2021alphafold,lin2023evolutionary} and materials property optimization~\citep{merchant2023scaling}. A key enabler of this progress is the emergence of Large Language Models (LLMs), which provide strong capabilities in natural language understanding, knowledge representation, and multi-step reasoning, enabling applications such as scientific literature analysis, hypothesis generation~\citep{kumbhar2025hypothesis}, and experimental planning~\citep{boiko2023autonomous}. Despite these successes, most existing LLMs are trained predominantly on general-domain corpora~\citep{taylor2022galactica}, resulting in limited coverage of specialized scientific knowledge. This limitation is particularly pronounced in engineering sciences, where problem-solving is governed by strict physical constraints and complex multi-physics interactions~\citep{raissi2019physics}.

Combustion science, a foundational discipline in energy systems and aerospace engineering, studies the coupled dynamics of chemical reactions, fluid transport, and energy conversion during fuel oxidation~\citep{wu2025physicsinformedmachinelearningcombustion}. From a modeling perspective, combustion problems exhibit several properties that pose substantial challenges for general-purpose LLMs. First, effective reasoning requires the integration of heterogeneous domain knowledge spanning chemical kinetics, fluid mechanics, and thermodynamics~\citep{ouyang2024structured}. Second, combustion processes are inherently spatiotemporal and nonlinear, involving cross-scale interactions from microscopic reaction pathways to macroscopic flow structures. Third, valid reasoning is constrained by fundamental physical laws, including mass, momentum, and energy conservation, violations of which render predictions physically meaningless~\citep{baez2024guaranteeing}. Consequently, the primary challenge in combustion-related tasks lies not in static state prediction, but in process-level reasoning under explicit physical constraints.

Current LLMs face two major bottlenecks when applied to combustion science. The first is a domain knowledge deficit: combustion-specific concepts, equations, and modeling paradigms are sparsely represented in general training corpora, limiting the model's ability to interpret specialized symbolic systems such as reaction mechanisms and turbulent combustion models~\citep{bran2024chemcrow,zhao2025chemdfm}. The second is the lack of physicochemical consistency in reasoning. Without explicit awareness of conservation laws and kinetic constraints, LLMs are prone to generating physically implausible outputs when confronted with tightly coupled multi-parameter scenarios, such as reaction pathways that violate chemical kinetics or efficiency estimates that contradict energy conservation. While retrieval-augmented generation (RAG) can partially mitigate knowledge sparsity by injecting external documents at inference time~\citep{Lewis2020Retrieval,zhong2025benchmarking}, it does not address inconsistencies arising during multi-step reasoning. Existing domain adaptation approaches, which typically focus on a single training stage, therefore remain insufficient for robust deployment in combustion-related applications.

To address these challenges, we propose a full-stack domain-enhanced LLM workflow tailored for combustion science. The proposed framework jointly improves domain knowledge acquisition and physically consistent reasoning through dedicated corpus construction, multi-stage model optimization, and standardized evaluation. Our contributions are summarized as follows:
\begin{itemize}
    \item Construction of a large-scale, combustion-specific corpus derived from hundreds of thousands of relevant scientific publications, providing high-quality training data for domain-specific language models.
    \item Development of a full-stack domain adaptation workflow that integrates data generation, multi-stage optimization, and objective evaluation—encompassing incremental pre-training, supervised fine-tuning (SFT), reinforcement learning with verifiable rewards (RLVR), and the creation of a benchmark dataset (FlameBench). This workflow simultaneously addresses critical knowledge gaps and establishes a standardized basis for evaluating domain-specific model performance.
\end{itemize}

\section{Related Work}
\subsection{Domain-Specific LLMs in AI4S}
Large Language Models have recently been adapted to a range of scientific domains under the AI for Science (AI4S) paradigm. In bioinformatics, the AlphaFold series demonstrates that incorporating domain structure can enable accurate protein structure prediction \citep{jumper2021alphafold}. In the biomedical domain, models such as BioGPT leverage large-scale medical corpora to support tasks including question answering and document summarization \citep{peng2023biogpt}. Similar trends have emerged in materials science and molecular design, where domain-specialized approaches are trained on curated literature and experimental data to support composition design and property prediction \citep{GomezBombarelli2018Chemical}. These efforts collectively suggest that domain-specific data and inductive biases are essential for extending the capabilities of LLMs to scientific problem settings. However, despite the central role of combustion science in energy and aerospace applications, the development of domain-specific foundation models for combustion remains largely unexplored.

\subsection{Domain Adaptation Methods for LLMs}
A variety of strategies have been proposed to adapt general-purpose LLMs to specialized domains. Continued or incremental pre-training on domain-specific corpora has been shown to improve vocabulary coverage and factual knowledge acquisition \citep{Gururangan2020Pretraining}. Supervised fine-tuning (SFT) further aligns models with downstream tasks through instruction-style datasets \citep{Wei2021Finetuned}. Retrieval-augmented generation (RAG) augments model outputs with external knowledge sources, mitigating factual hallucinations in knowledge-intensive settings. In addition, post-training alignment methods, including Direct Preference Optimization (DPO) \citep{Rafailov2023DPO} and Reinforcement Learning from Human Feedback (RLHF) \citep{Ouyang2022RLHF}, aim to improve output consistency and preference alignment. While effective in general domains, these approaches exhibit limitations in combustion-related applications: incremental pre-training alone does not enforce physical consistency, SFT provides limited supervision for tightly coupled dynamical reasoning, RAG struggles with integrated multi-physics reasoning, and RLHF is constrained by the scalability of expert feedback in highly specialized scientific domains.

\subsection{AI Applications in Combustion Science}
Machine learning has long been extensively applied to combustion-related research \citep{wu2025physicsinformedmachinelearningcombustion}, and it has yielded remarkable results particularly in the aspects of predictive modeling for combustion parameters and reduced-order characterization of complex physical processes. In recent years, with the advancement of large language models (LLMs), the academic community has begun to explore their novel applications in combustion science—for instance, constructing domain-adaptive solutions via retrieval-augmented generation (RAG) \citep{sharma2024reliable}. Such explorations hold certain value, yet they remain overall in the proof-of-concept stage: suffering from a narrow knowledge coverage and weak scene generalization ability, these approaches not only cannot avoid factual hallucinations but also fail to support practical engineering deployment. 
\section{Methodology}

\subsection{Overview}
We propose a full-stack domain-enhanced workflow that adapts general-purpose LLMs to combustion science by jointly addressing domain knowledge acquisition and physically consistent reasoning. The workflow consists of four components: (i) construction of a large-scale combustion-specific corpus, (ii) multi-stage model adaptation, (iii) reinforcement learning, and (iv) standardized evaluation using a domain-specific benchmark. Starting from a general base model, the framework progressively injects domain knowledge and enforces physical consistency through incremental pre-training, supervised fine-tuning, and reinforcement learning, followed by systematic evaluation on \textsc{FlameBench}.

\subsection{Corpus Construction}

\subsubsection{Data Sources}
To balance domain specialization and general reasoning capability, we construct a mixed corpus of approximately 30B tokens. The corpus includes 5B tokens of combustion-related data and 25B tokens of general pre-training data. The domain corpus is composed of English and Chinese academic publications in combustion science, as well as curated scientific encyclopedic resources covering physics, chemistry, and engineering fundamentals. The general corpus is sampled from open-source mid-training datasets \citep{Olmo2025Olmo3,Soldaini2024Dolma}, ensuring retention of general language understanding and commonsense reasoning.
\begin{figure}[!htbp]  
    \centering
    \includegraphics[width=\linewidth]{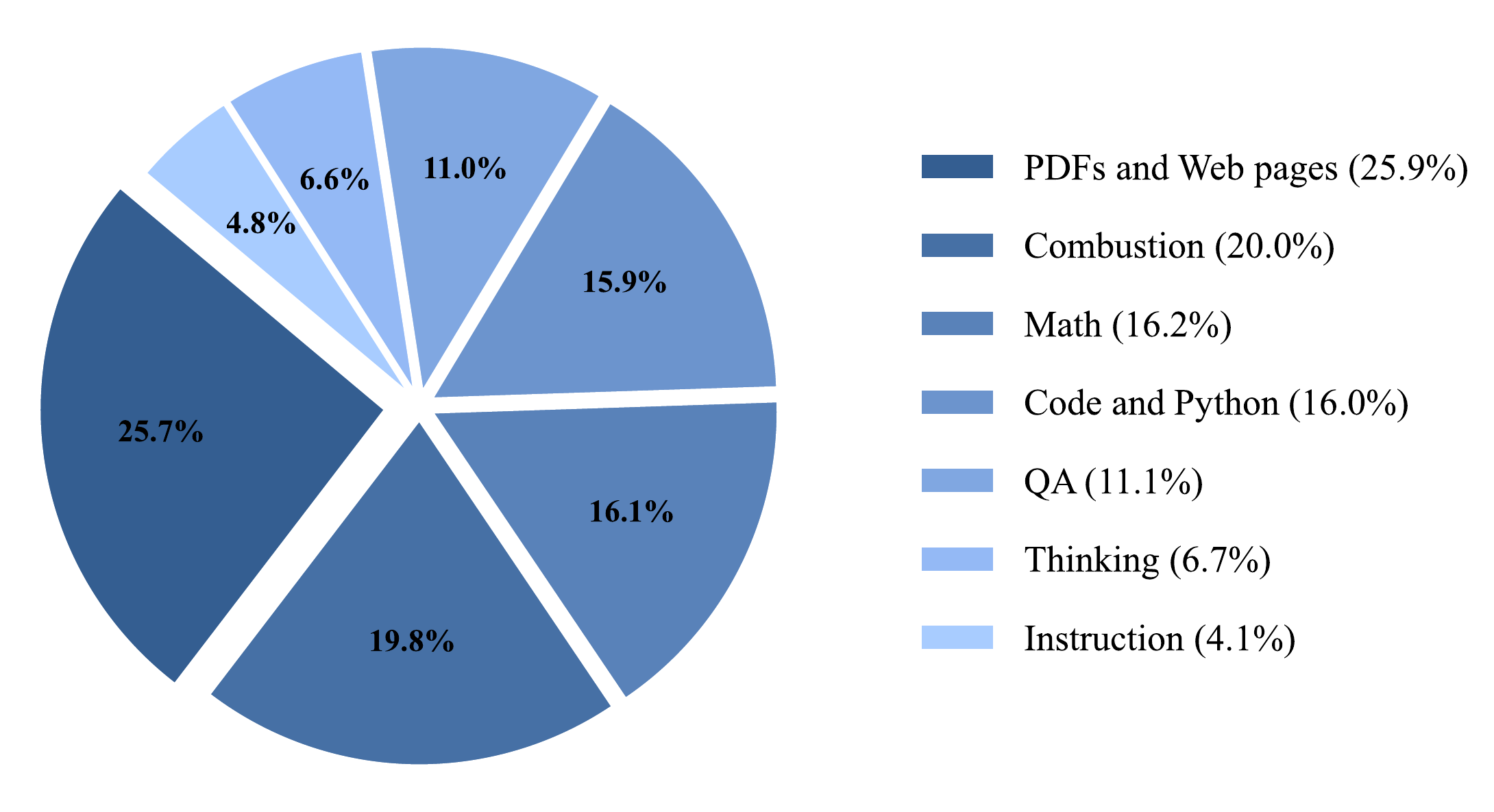}
    \caption{Dataset token distribution by category. }
    \label{fig:token_distribution}
    \vspace{-0.5em}  
\end{figure}
\subsubsection{Automated Processing Pipeline}
Raw documents are transformed into structured training data through an automated pipeline. PDF documents are first parsed and converted into Markdown format\citep{Wang2024MinerU,Wei2025DeepSeekOCR} . Multi-level deduplication is then performed at both exact and approximate levels, using hashing for exact matching and MinHash-based similarity detection for approximate matching \citep{Jennings2023NeMoCurator}. Quality control is performed through a combination of rule-based filtering and model-based evaluation, where low-quality fragments are identified using perplexity and relevance scoring and subsequently repaired or removed. This process yields a high-quality structured corpus suitable for large-scale model training.
\begin{figure}[!htbp] 
    \centering
    \includegraphics[width=0.9\linewidth]{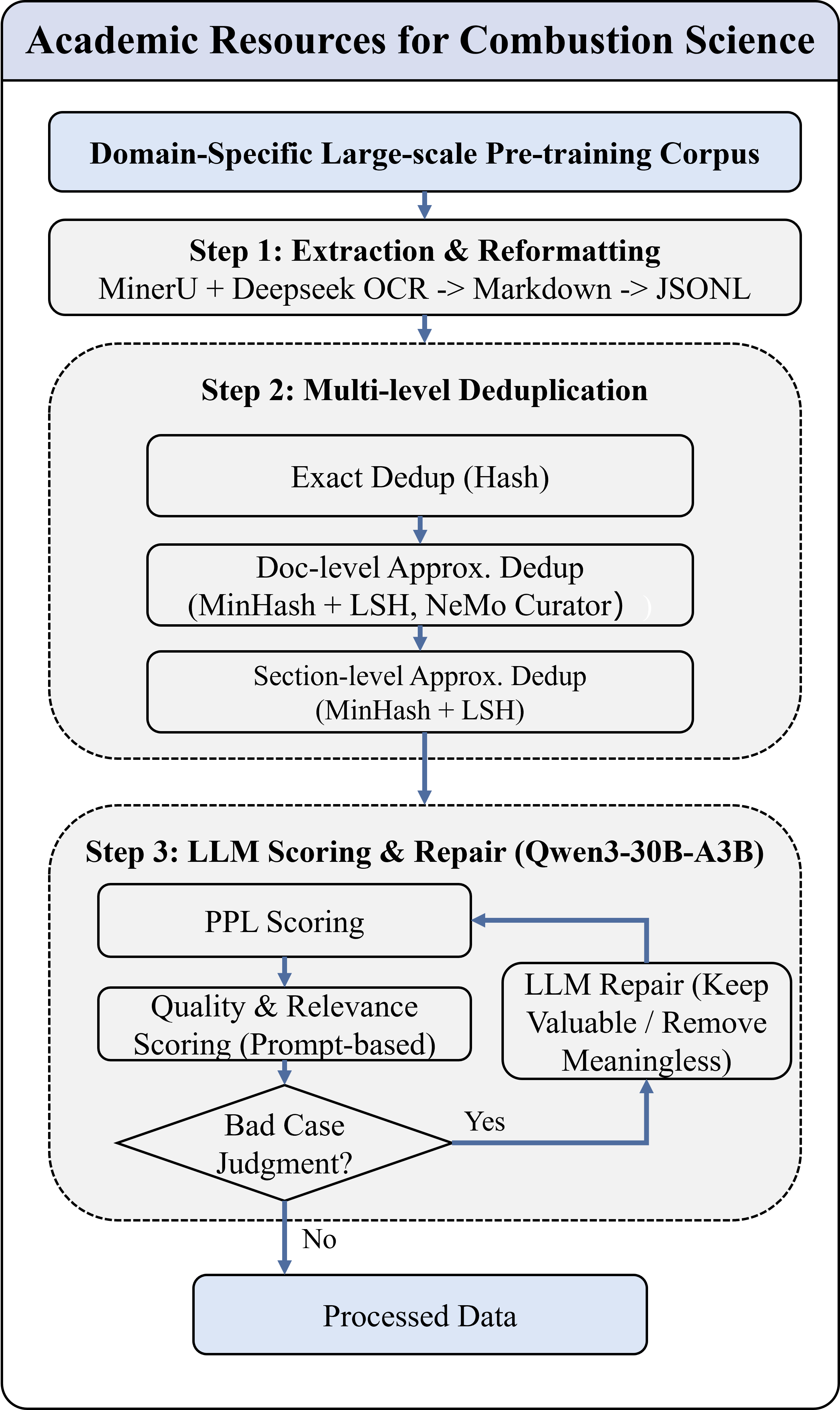} 
    \caption{Data processing pipeline for the combustion-specific pre-training corpus.}
    \label{fig:data_flow}
      \vspace{-0.5em} 
\end{figure}
\subsubsection{Post-training Data Construction}
Based on the curated corpus, we construct datasets for supervised fine-tuning (SFT) and reinforcement learning with verifiable rewards (RLVR). The SFT dataset includes 800K general instruction-following samples \citep{Olmo2025Olmo3} and 12K combustion-specific chain-of-thought examples spanning knowledge queries, formula derivation, experimental analysis, and literature summarization. The RLVR dataset consists of 7K complex reasoning instances targeting multi-parameter coupling and process-level deduction in combustion scenarios.

\subsection{Multi-Stage Model Adaptation}

\subsubsection{Continue Pre-training}
We perform Continue pre-training(CPT) on the mixed corpus to inject combustion-specific knowledge while preserving general language capabilities\citep{Ibrahim2024Continual,Que2024DCPT}. Training is conducted for one epoch using a conservative learning rate schedule to avoid catastrophic forgetting. This stage enables the model to acquire domain terminology, core concepts, and fundamental physical relationships relevant to combustion science.
\begin{figure}[!htbp]  
    \centering
    \includegraphics[width=\linewidth]{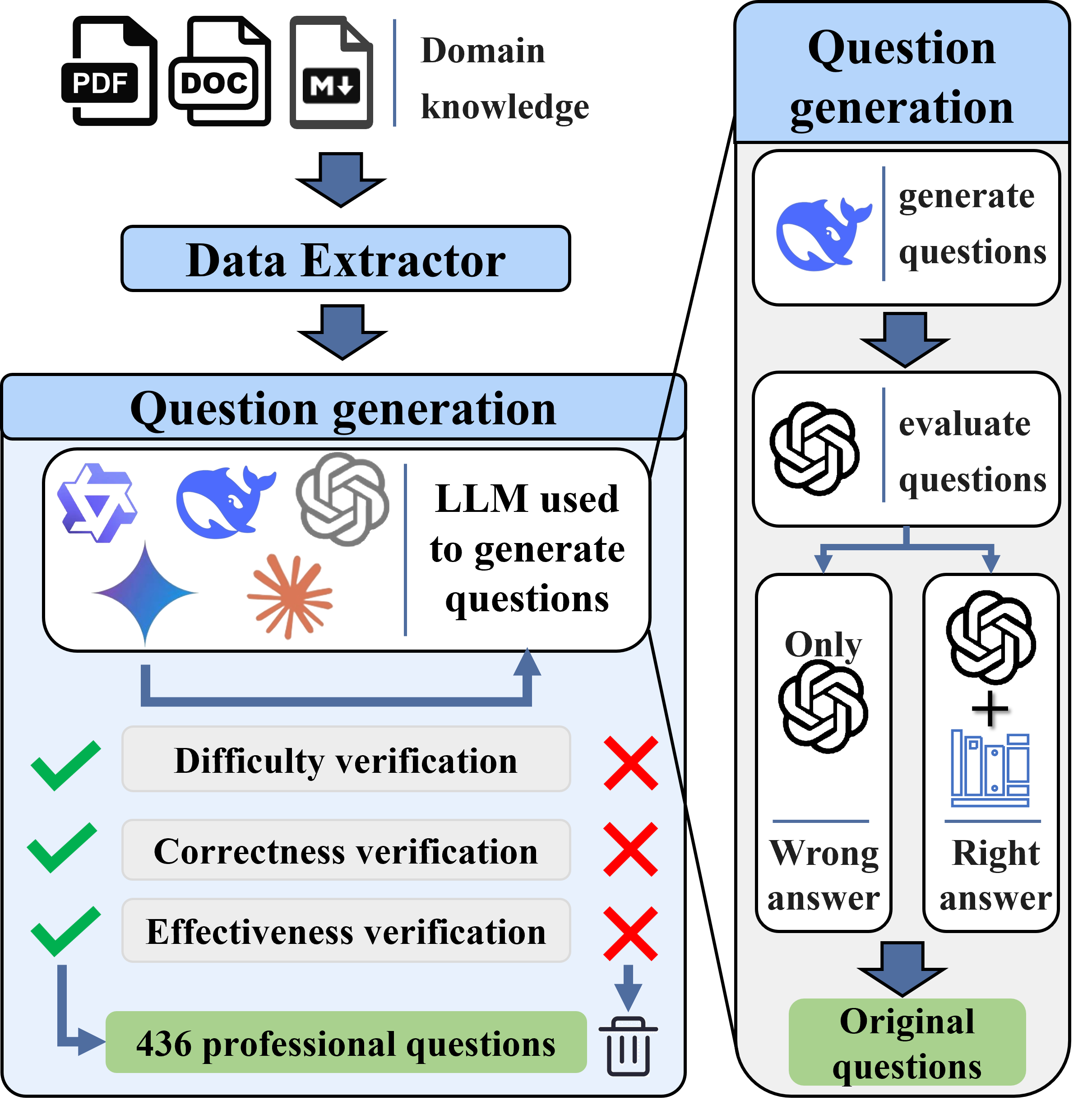}
    \caption{The construction pipeline of FlameBench.}
    \label{fig:flamebench}
    \vspace{-0.5em}  
\end{figure}
\subsubsection{Supervised Fine-Tuning}
Following CPT, we apply supervised fine-tuning in two phases. The model is first aligned with general instruction-following behavior using a large-scale open-source instruction dataset. It is then adapted to combustion-specific tasks through fine-tuning on domain-specific chain-of-thought data. Optimization is performed using cross-entropy loss with a cosine learning rate schedule.

\subsubsection{Reinforcement Learning}
To further improve reasoning reliability under physical constraints, we introduce reinforcement learning with verifiable rewards (RLVR) based on the GRPO framework\citep{deepseekr1}. A binary reward function is defined to explicitly penalize violations of domain knowledge and physical consistency. We regularize policy updates using a KL divergence constraint to maintain training stability. This stage enables the model to generate physically plausible and logically consistent solutions for complex combustion reasoning tasks.

\subsection{FlameBench: A Domain-Specific Evaluation Benchmark}
To evaluate domain knowledge retention and constrained reasoning, we introduce FlameBench, a benchmark tailored to combustion science. The benchmark is constructed from high-information-density fragments extracted from peer-reviewed literature, dissertations, and domain-specific code repositories. Questions are generated and verified through an automated pipeline with dual validation, followed by expert refinement. The final benchmark consists of 436 high-quality questions covering eight combustion subfields, with each question grounded in a unique source reference to ensure reproducibility and verifiability. Fig.~3 illustrates the construction workflow of FlameBench.

\section{Experiments and Result Analysis}
\label{sec:experiments}

\subsection{Experimental Setup}

\subsubsection{Base Model and Hardware Environment}
We adopt Qwen-8B \citep{Qwen2025Qwen3} as the base model, chosen for its favorable trade-off between general-language capability and training efficiency. Continued pre-training (CPT), supervised fine-tuning (SFT), and evaluation were conducted on a distributed cluster equipped with 80 Huawei Ascend 910B accelerators. The reinforcement learning with verifiable rewards (RLVR) stage was executed on a separate node with 8 NVIDIA A800 GPUs.

\subsubsection{Training Stages and Control Groups}
The primary experimental variable is the training stage. To quantify the incremental contribution of each component in our pipeline, we consider the following model variants:
\begin{itemize}
\item \textbf{Baseline}: The original Qwen-8B model without domain adaptation.
\item \textbf{CPT}: Qwen-8B is further trained on a 30B-token hybrid corpus (5B combustion-domain tokens and 25B general-domain tokens).
\item \textbf{SFT-General}: The CPT model fine-tuned on 800K general-purpose instruction-following samples.
\item \textbf{SFT-Combustion}: The SFT-General model further fine-tuned on 12K combustion-specific chain-of-thought (CoT) samples, targeting domain-specific reasoning tasks , and serving as the cold-start model for subsequent RLVR optimization.
\item \textbf{RLVR-Opt}: The SFT-Combustion model further trained with RLVR on 7K complex samples.
\item \textbf{RAG-Methods}: As illustrated in Figure 4, the RAG system is implemented using FAISS for indexing, BGE-M3 \citep{Chen2024BGE} for embedding generation, and LangChain 
\citep{LangChain2022} for orchestration.

\begin{figure}[htbp]  
    \centering
    \includegraphics[width=0.9\linewidth]{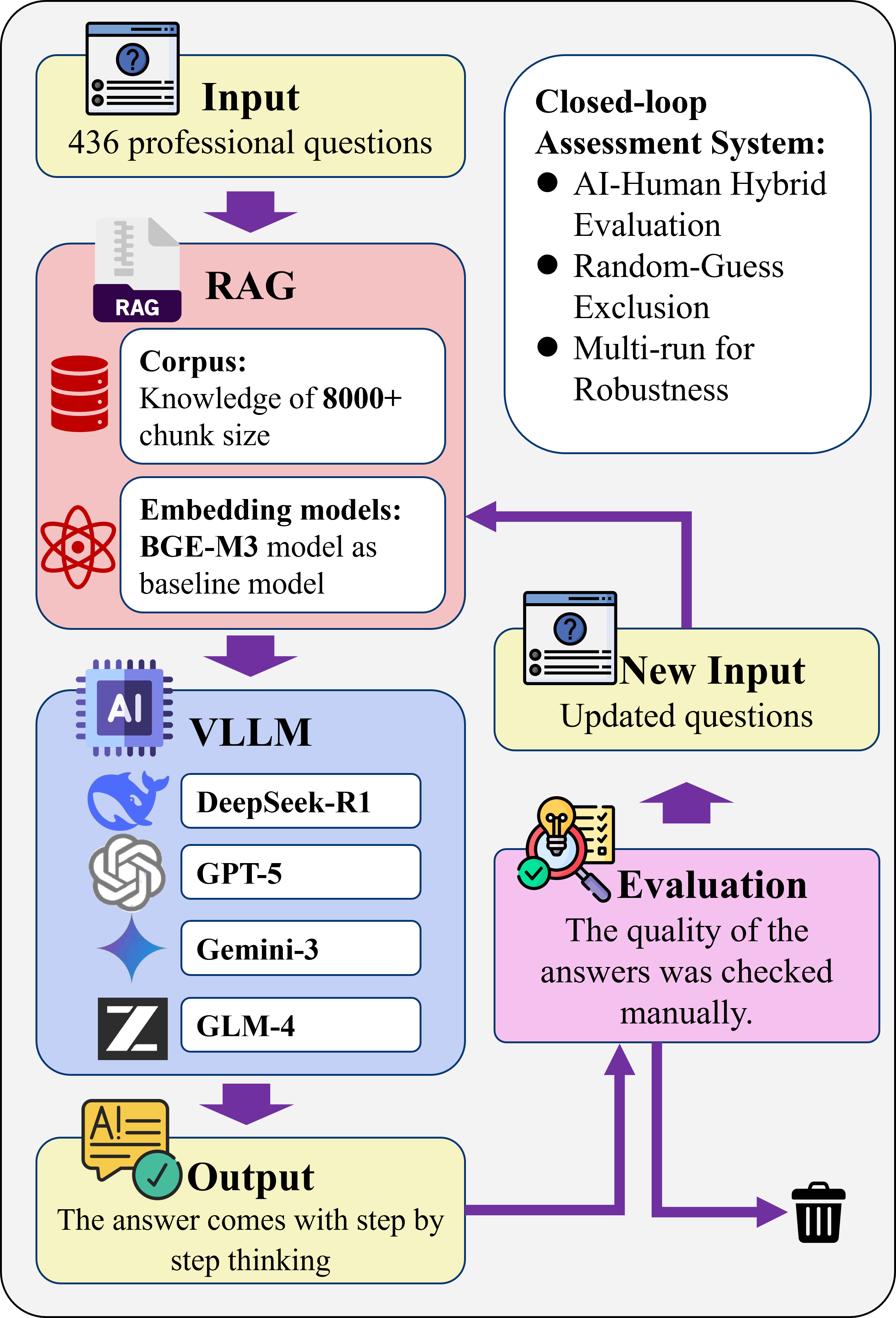}
    \caption{Architecture of the Closed-loop RAG Assessment System.}
    \label{fig:rag}
    \vspace{-0.5em}  
\end{figure}
\end{itemize}
\begin{figure*}[t]  
    \centering
    \includegraphics[width=1\textwidth]{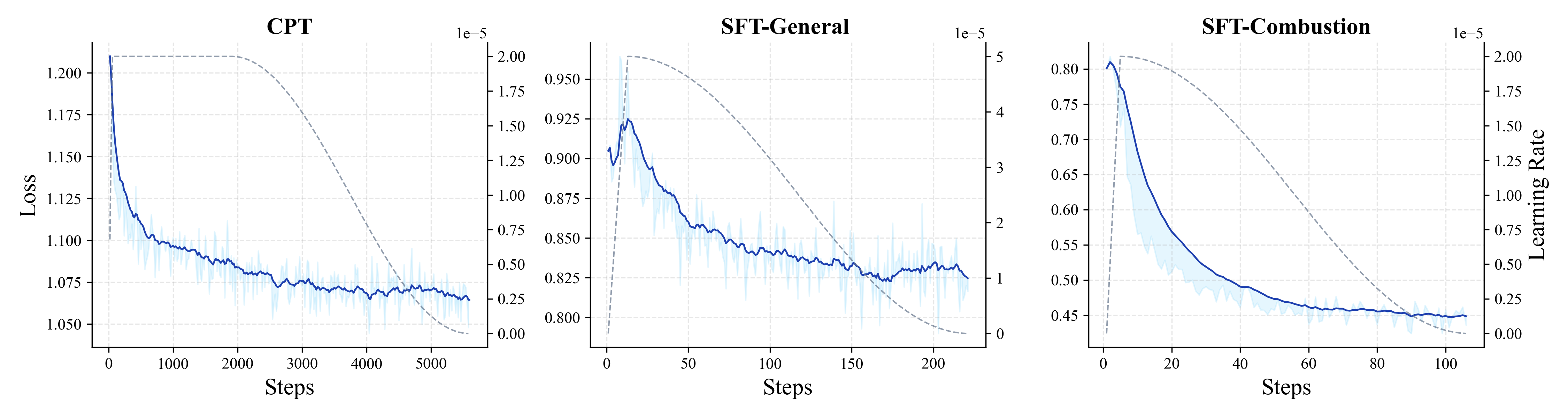} 
    \caption{Loss and learning rate curves during the Continue Pre-training and SFT stages.}
    \label{fig:loss-lr-curves}
\end{figure*}

\subsubsection{Evaluation Protocol}
We evaluate all models on FlameBench, a domain-specific benchmark designed to assess reasoning in combustion science. Performance is measured using multiple-choice accuracy, which directly reflects the model’s ability to apply fundamental principles such as chemical kinetics, transport phenomena, and multi-physics coupling. All evaluation items undergo a three-stage validation process—difficulty filtering, correctness verification, and expert calibration—to ensure coverage, reliability, and reproducibility.

\subsection{Results and Analysis}

\subsubsection{Performance across Training Stages}
Table~\ref{tab:performance} reports the performance of different training stages on FlameBench. We observe a clear monotonic improvement as the training pipeline progresses, indicating that each stage contributes complementary benefits while maintaining practical inference efficiency.

\begin{table}[htbp]
\caption{Performance Comparison of Multi-stage Training}
\label{tab:performance}
\centering
\begin{tabular}{lc}
\toprule
\textbf{Model Group} & \textbf{ Accuracy (\%)} \\
\midrule
Qwen3-8B-Base & 26.8 \\
CPT & 33.3 \\
SFT-General & 33.5 \\
SFT-Combustion & 35.1 \\
\textbf{RLVR-Opt} & \textbf{43.8} \\
\bottomrule
\end{tabular}
\end{table}

\begin{table}[htbp]
\caption{Performance Comparison of Different Models}
\label{tab:model-performance}
\centering
\begin{tabular}{lr}
\toprule
\textbf{Model} & \textbf{Accuracy (\%)} \\
\midrule
GPT-5          & 15.60 \\
GLM-4          & 32.64 \\
Gemini Pro     & 32.10 \\
DeepSeek-R1    & 28.37 \\
Average Score & \textbf{27.18} \\
\bottomrule
\end{tabular}
\end{table}

\begin{table}[htbp]
\caption{Performance of RAG with Different Models}
\label{tab:rag-performance}
\centering
\begin{tabular}{lr}
\toprule
\textbf{Method} & \textbf{Accuracy (\%)} \\
\midrule
RAG + GPT-5          & 16.52 \\
RAG + GLM-4          & 32.09\\
RAG + Gemini Pro     & 27.80 \\
RAG + DeepSeek-R1    & 28.54 \\
Average Score & \textbf{26.24} \\
\bottomrule
\end{tabular}
\end{table}
\paragraph{Impact of Continued Pre-training}
Relative to the baseline (Qwen3-8B-Base, accuracy = 26.8\%), Continued Pre-training (CPT) yields a substantial improvement of 6.5 percentage points (accuracy = 33.3\%), confirming that large-scale hybrid-corpus pre-training effectively injects foundational combustion knowledge. Notably, CPT outperforms the average accuracy of RAG-based methods (26.24\%) by 7.06 percentage points, and even surpasses the best-performing RAG model (RAG + GLM-4, 32.09\%). This result demonstrates that internalized domain knowledge provides a stronger inductive bias for combustion reasoning than external retrieval, which often suffers from irrelevant context injection and limited multi-step reasoning capability.

\paragraph{Effects of Supervised Fine-tuning}
General-purpose SFT (SFT-General, 33.5\%) only brings marginal improvement over CPT, indicating that it primarily enhances instruction adherence rather than domain reasoning. In contrast, domain-specific SFT (SFT-Combustion, 35.1\%) further boosts accuracy by 1.8 percentage points, as it aligns the model with combustion-specific problem-solving patterns such as chemical kinetics derivation and thermodynamics constraint application. Specifically, SFT-Combustion outperforms closed-source models including DeepSeek-R1 (28.37\%) and Gemini Pro (32.10\%) on FlameBench, and approaches the performance of GLM-4 (32.64\%) in the turbulent combustion subfield. The modest gain also reflects the high complexity of FlameBench, where standard SFT alone is insufficient to address multi-physics coupling challenges.

\begin{figure}[tb]  
    \centering
    \begin{subfigure}[t]{0.5\textwidth}  
        \centering
        \includegraphics[width=\linewidth]{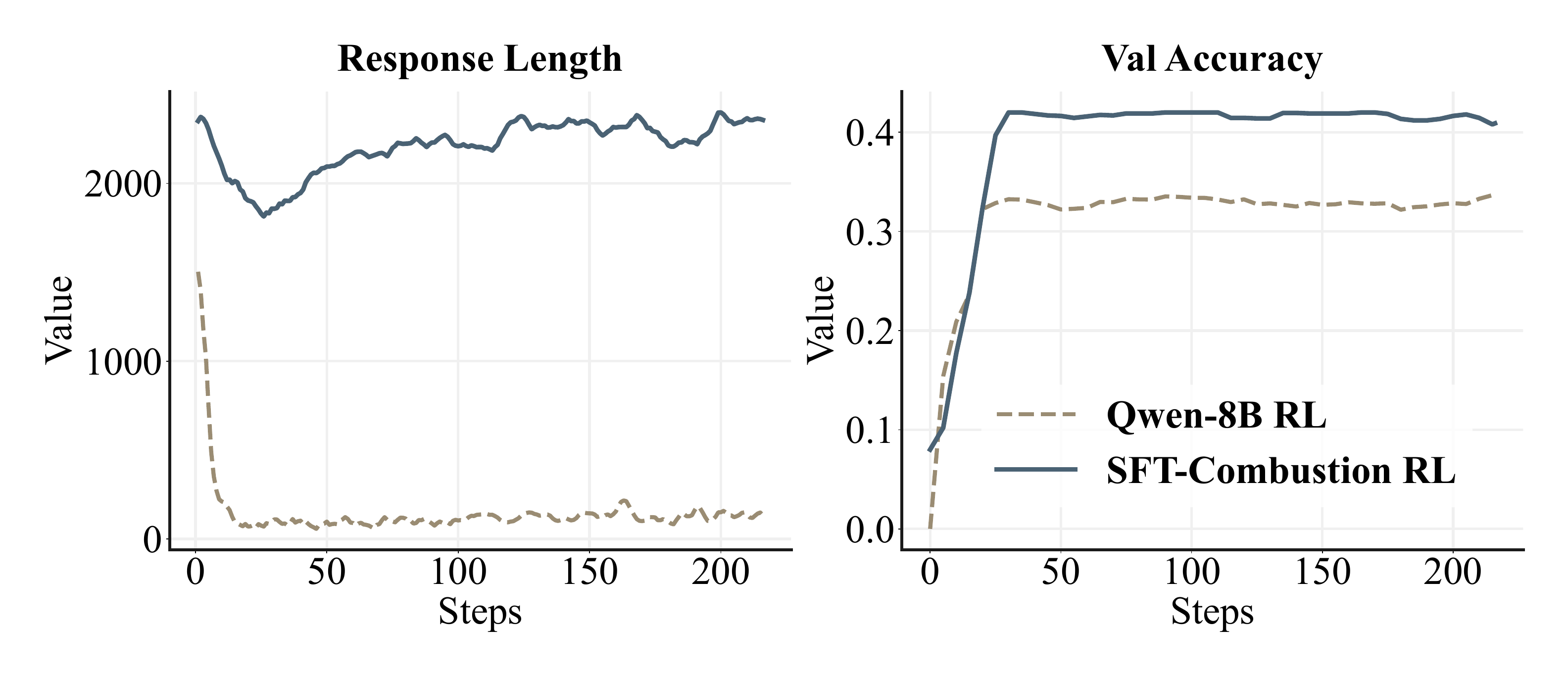}
        \caption{Response Length and Validation Accuracy}
        \label{fig:len_val}
    \end{subfigure}
    \hfill  
    \begin{subfigure}[t]{0.5\textwidth}
        \centering
        \includegraphics[width=\linewidth]{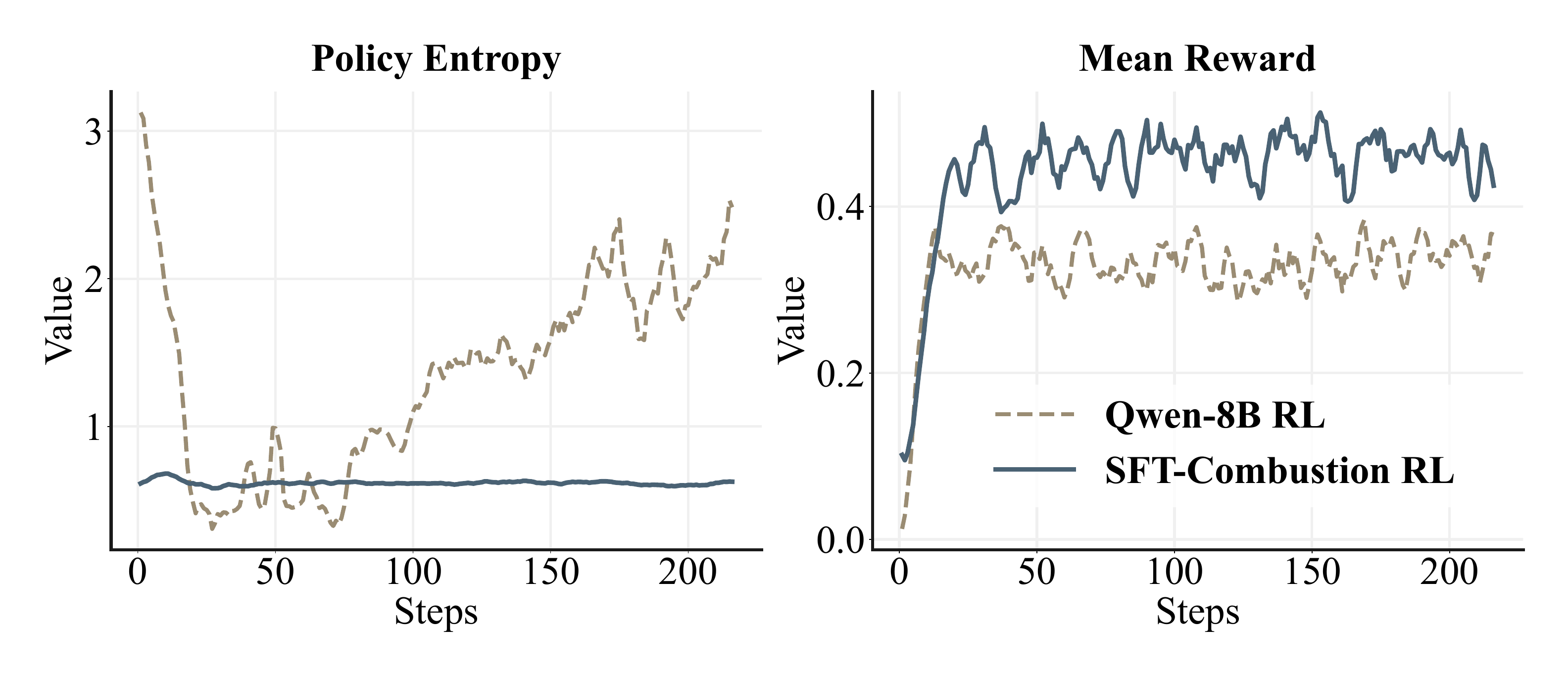}
        \caption{Mean Reward and Policy Entropy}
        \label{fig:reward_entropy}
        \vspace{-0.5em}  
    \end{subfigure}
    
    \vspace{0.5em}  
    \caption{Comparison of training metrics between SFT-Combustion RL and Qwen-8B RL.}
    \label{fig:training_metrics}
    \vspace{-1em}  
\end{figure}

\paragraph{Benefits of RLVR Optimization}
Reinforcement Learning from Verifiable Rewards (RLVR) substantially improves upon the SFT-Combustion baseline, increasing accuracy from 35.1\% to 43.8\%. As shown in Figure~\ref{fig:training_metrics}, the vanilla Qwen-8B RL baseline exhibits a rapid reduction in response length—from approximately 1,500 tokens to 100--200 tokens—reflecting a tendency toward shortcut-driven policy collapse during optimization. In contrast, the RLVR model initialized from SFT-Combustion maintains a stable response length of around 2,000 tokens throughout training.

This length stability, together with consistently higher validation accuracy and mean rewards under a low-entropy policy regime, is indicative of convergence toward a more deterministic and stable optimization regime, rather than reliance on exploitative or reward-hacking behaviors. While sustained output length alone does not guarantee improved reasoning fidelity, these results are consistent with more systematic and physically grounded reasoning processes. Overall, our findings suggest that RLVR, when fortified with domain-specific priors, stabilizes policy optimization and reduces physics-inconsistent generations in combustion science tasks.

\subsection{Comparison with RAG Methods}
A direct quantitative comparison between RLVR-Opt and RAG baselines (Table \ref{tab:rag-performance}) highlights three distinct advantages of our end-to-end training paradigm. First, knowledge internalization enables RLVR-Opt to outperform the best RAG model (RAG + GLM-4, 32.09\%) by 11.71 percentage points, avoiding the context interference and performance ceiling of retrieval-based systems. Second, domain-specific reasoning emerges from multi-stage optimization, allowing the model to handle cross-disciplinary coupling (e.g., fluid mechanics + chemical reactions) more reliably than RAG, which often struggles with integrating heterogeneous knowledge fragments. Third, inference efficiency is drastically improved: RLVR-Opt eliminates the overhead of document retrieval and embedding matching, reducing inference latency for interactive scientific analysis and real-time simulation.

\subsection{Experimental Conclusion}
Overall, the results demonstrate that the proposed CPT--SFT--RLVR framework substantially improves LLM performance in combustion science. Hybrid-corpus CPT is necessary for effective knowledge injection, while RLVR constitutes the key step for surpassing the limitations of both SFT and RAG by explicitly optimizing scientific reasoning. The resulting model achieves state-of-the-art performance on FlameBench while maintaining efficient end-to-end inference.

\section{Conclusion}
This work addresses the limitations of general-purpose large language models (LLMs) in combustion science, particularly the lack of domain-specific knowledge. We propose a full-stack domain-enhanced workflow that integrates three core components: construction of a dedicated combustion corpus, multi-stage model training (CPT, SFT, and RLVR), and a standardized evaluation benchmark (FlameBench).

Experimental results demonstrate that the proposed workflow systematically improves both domain knowledge retention and reasoning capabilities. Specifically, the model optimized through CPT--SFT--RLVR achieves the highest performance on FlameBench, outperforming general-purpose LLMs and retrieval-augmented baselines. 

Beyond combustion science, this workflow provides a generalizable paradigm for enhancing LLMs in engineering domains characterized by strict physical laws and multidisciplinary coupling. Future work will explore extending this approach to broader AI-for-Science applications and more complex experimental scenarios, aiming to further integrate LLMs into scientific discovery workflows.

\section*{Impact Statement}
This paper presents work whose goal is to advance the field of Machine Learning. There are many potential societal consequences of our work, none which we feel must be specifically highlighted here.
\bibliography{myrefs}

@article{jumper2021alphafold,
  title={Highly accurate protein structure prediction with {AlphaFold}},
  author={Jumper, John and Evans, Richard and Pritzel, Alexander and Green, Tim and Figurnov, Michael and Ronneberger, Olaf and Tunyasuvunakool, Kathryn and Bates, Russ and {\v Z}{\'\i}dek, Augustin and Potapenko, Anna and Bridgland, Alex and others},
  journal={Nature},
  volume={596},
  number={7873},
  pages={583--589},
  year={2021},
  publisher={Nature Publishing Group},
  doi={10.1038/s41586-021-03819-2},
  url={https://www.nature.com/articles/s41586-021-03819-2}
}

@article{lin2023evolutionary,
  title={Evolutionary-scale prediction of atomic-level protein structure with a language model},
  author={Lin, Zeming and Akin, Halil and Rao, Roshan and Hie, Brian and Zhu, Zhongkai and Lu, Wenting and Smetanin, Nikita and Verkuil, Robert and Kabeli, Ori and Shmueli, Yaniv and dos Santos Costa, Allan and others},
  journal={Science},
  volume={379},
  number={6637},
  pages={1123--1130},
  year={2023},
  publisher={American Association for the Advancement of Science},
  doi={10.1126/science.ade2574},
  url={https://www.science.org/doi/10.1126/science.ade2574}
}

@article{merchant2023scaling,
  title={Scaling deep learning for materials discovery},
  author={Merchant, Amil and Batzner, Simon and Schoenholz, Samuel S and Aykol, Muratahan and Cheon, Gowoon and Cubuk, Ekin Dogus},
  journal={Nature},
  volume={624},
  number={7990},
  pages={80--85},
  year={2023},
  publisher={Nature Publishing Group},
  doi={10.1038/s41586-023-06735-9},
  url={https://www.nature.com/articles/s41586-023-06735-9}
}

@misc{kumbhar2025hypothesis,
  title={Hypothesis Generation for Materials Discovery and Design Using Goal-Driven and Constraint-Guided LLM Agents},
  author={Shrinidhi Kumbhar and Venkatesh Mishra and Kevin Coutinho and Divij Handa and Ashif Iquebal and Chitta Baral},
  year={2025},
  eprint={2501.13299},
  archivePrefix={arXiv},
  primaryClass={cs.CL},
  url={https://arxiv.org/abs/2501.13299}
}

@article{boiko2023autonomous,
  title={Autonomous chemical research with large language models},
  author={Boiko, Daniil A and MacKnight, Robert and Kline, Ben and Gomes, Gabriel},
  journal={Nature},
  volume={624},
  number={7992},
  pages={570--578},
  year={2023},
  publisher={Nature Publishing Group},
  doi={10.1038/s41586-023-06792-0},
  url={https://www.nature.com/articles/s41586-023-06792-0}
}

@misc{taylor2022galactica,
  title={Galactica: A Large Language Model for Science},
  author={Ross Taylor and Marcin Kardas and Guillem Cucurull and Thomas Scialom and Anthony Hartshorn and Elvis Saravia and Andrew Poulton and Viktor Kerkez and Robert Stojnic},
  year={2022},
  eprint={2211.09085},
  archivePrefix={arXiv},
  primaryClass={cs.CL},
  url={https://arxiv.org/abs/2211.09085}
}

@article{raissi2019physics,
  title={Physics-informed neural networks: A deep learning framework for solving forward and inverse problems involving nonlinear partial differential equations},
  author={Raissi, Maziar and Perdikaris, Paris and Karniadakis, George E},
  journal={Journal of Computational Physics},
  volume={378},
  pages={686--707},
  year={2019},
  publisher={Elsevier},
  doi={10.1016/j.jcp.2018.10.045},
  url={https://www.sciencedirect.com/science/article/pii/S0021999118307125}
}

@misc{wu2025physicsinformedmachinelearningcombustion,
  title={Physics-informed machine learning for combustion: A review},
  author={Jiahao Wu and Xutun Wang and Guihua Zhang and Jiayue Liu and Xin Li and Yang Zhang and Hai Zhang and Junfu Lyu and Bing Wang and Yuxin Wu},
  year={2025},
  eprint={2509.03347},
  archivePrefix={arXiv},
  primaryClass={physics.chem-ph},
  url={https://arxiv.org/abs/2509.03347}
}

@article{ouyang2024structured,
  title={Structured chemistry reasoning with large language models},
  author={Ouyang, Siru and Zhang, Zhuosheng and Yan, Bing and Liu, Xuan and Choi, Yejin and Han, Jiawei and Qin, Lianhui},
  journal={Conference on Machine Learning},
  year={2024}
}

@article{baez2024guaranteeing,
  title={Guaranteeing conservation laws with projection in physics-informed neural networks},
  author={Baez, Anthony and Zhang, Wang and Ma, Ziwen and Das, Subhro and Nguyen, Lam M and Daniel, Luca},
  journal={arXiv preprint arXiv:2410.17445},
  year={2024},
  url={https://arxiv.org/abs/2410.17445}
}

@article{bran2024chemcrow,
  title={Augmenting large language models with chemistry tools},
  author={M. Bran, Andres and Cox, Sam and Schilter, Oliver and Baldassari, Carlo and White, Andrew D. and Schwaller, Philippe},
  year={2024},
  month={05},
  journal={Nature Machine Intelligence},
  volume={6},
  number={5},
  pages={525--535},
  issn={2522-5839},
  doi={10.1038/s42256-024-00832-8},
  url={https://doi.org/10.1038/s42256-024-00832-8}
}

@article{zhao2025chemdfm,
  title={Developing ChemDFM as a large language foundation model for chemistry},
  author={Zhao, Zihan and Ma, Da and Chen, Lu and Sun, Liangtai and Li, Zihao and Xia, Yi and Chen, Bo and Xu, Hongshen and Zhu, Zichen and Zhu, Su and Fan, Shuai and Shen, Guodong and Yu, Kai and Chen, Xin},
  year={2025},
  month={04},
  journal={Cell Reports Physical Science},
  volume={6},
  number={4},
  publisher={Elsevier},
  issn={2666-3864},
  doi={10.1016/j.xcrp.2025.102523},
  url={https://doi.org/10.1016/j.xcrp.2025.102523}
}

@InProceedings{Lewis2020Retrieval,
  author={P. Lewis and others},
  title={Retrieval-augmented generation for knowledge-intensive NLP tasks},
  booktitle={NeurIPS},
  year={2020}
}

@article{zhong2025benchmarking,
  title={Benchmarking Retrieval-Augmented Generation for Chemistry},
  author={Xianrui Zhong and Bowen Jin and Siru Ouyang and Yanzhen Shen and Qiao Jin and Yin Fang and Zhiyong Lu and Jiawei Han},
  year={2025},
  eprint={2505.07671},
  archivePrefix={arXiv},
  primaryClass={cs.CL},
  url={https://arxiv.org/abs/2505.07671}
}

@article{peng2023biogpt,
  author={Peng, Yu and Wang, Zhihao and Zhang, Cheng and Li, Haotian and Liu, Jing and Xu, Jianzhu},
  title={{BioGPT}: Generative pre-trained transformer for biomedical text generation and mining},
  journal={Bioinformatics},
  year={2023},
  volume={39},
  number={5},
  pages={btad325},
  doi={10.1093/bioinformatics/btad325}
}

@article{GomezBombarelli2018Chemical,
  author={G{\'o}mez-Bombarelli, Rafael and Duvenaud, David and S{\'a}nchez-Lengeling, Benjam{\'i}n and Sheberla, Doreen and Aguilera-Iparraguirre, Jorge and Hirzel, Timothy D. and Adams, Ryan P. and Aspuru-Guzik, Al{\'a}n},
  title={Automatic chemical design using a data-driven continuous representation of molecules},
  journal={ACS Central Science},
  year={2018},
  volume={4},
  number={2},
  pages={268--276},
  doi={10.1021/acscentsci.7b00572}
}

@InProceedings{Gururangan2020Pretraining,
  author={S. Gururangan and others},
  title={Don't stop pretraining: Adapt language models to domains and tasks},
  booktitle={ACL},
  year={2020}
}

@InProceedings{Wei2021Finetuned,
  author={J. Wei and others},
  title={Finetuned language models are zero-shot learners},
  booktitle={International Conference on Machine Learning},
  year={2021}
}

@InProceedings{Rafailov2023DPO,
  author={R. Rafailov and others},
  title={Direct preference optimization: Your language model is secretly a reward model},
  booktitle={NeurIPS},
  year={2023}
}

@InProceedings{Ouyang2022RLHF,
  author={L. Ouyang and others},
  title={Training language models to follow instructions with human feedback},
  booktitle={NeurIPS},
  year={2022}
}

@article{sharma2024reliable,
  title={A reliable knowledge processing framework for combustion science using foundation models},
  author={Sharma, Vansh and Raman, Venkat},
  journal={Energy and AI},
  volume={16},
  pages={100365},
  year={2024},
  publisher={Elsevier}
}

@Article{Olmo2025Olmo3,
  author={{Team Olmo} and others},
  title={Olmo 3},
  journal={arXiv preprint arXiv:2512.13961},
  year={2025},
  url={https://arxiv.org/abs/2512.13961}
}

@Article{Soldaini2024Dolma,
  author={Luca Soldaini and others},
  title={Dolma: An Open Corpus of Three Trillion Tokens for Language Model Pretraining Research},
  journal={arXiv preprint arXiv:2402.00159},
  year={2024},
  url={https://arxiv.org/abs/2402.00159}
}

@Article{Wang2024MinerU,
  author={Bin Wang and others},
  title={MinerU: An Open-Source Solution for Precise Document Content Extraction},
  journal={arXiv preprint arXiv:2409.18839},
  year={2024},
  url={https://arxiv.org/abs/2409.18839}
}

@Article{Wei2025DeepSeekOCR,
  author={Haoran Wei and others},
  title={DeepSeek-OCR: Contexts Optical Compression},
  journal={arXiv preprint arXiv:2510.18234},
  year={2025},
  url={https://arxiv.org/abs/2510.18234}
}

@Misc{Jennings2023NeMoCurator,
  author={Joseph Jennings and others},
  title={Curating Trillion-Token Datasets: Introducing {NVIDIA} {NeMo} Data Curator},
  howpublished={NVIDIA Developer Blog},
  year={2023},
  month={Aug},
  url={https://developer.nvidia.com/blog/curating-trillion-token-datasets-introducing-nemo-data-curator/}
}

@Article{Ibrahim2024Continual,
  author={Adam Ibrahim and others},
  title={Simple and Scalable Strategies to Continually Pre-train Large Language Models},
  journal={arXiv preprint arXiv:2312.06946},
  year={2023},
  url={https://arxiv.org/abs/2312.06946}
}

@Article{Que2024DCPT,
  author={Haoran Que and others},
  title={D-CPT Law: Domain-Specific Continual Pre-Training Scaling Law for Large Language Models},
  journal={arXiv preprint arXiv:2406.01375},
  year={2024},
  url={https://arxiv.org/abs/2406.01375}
}

@Article{Qwen2025Qwen3,
  author={{Qwen Team} and others},
  title={Qwen3 Technical Report},
  journal={arXiv preprint arXiv:2505.09388},
  year={2025},
  url={https://arxiv.org/abs/2505.09388}
}

@article{deepseekr1,
  title={DeepSeek-R1: Incentivizing Reasoning Capability in LLMs via Reinforcement Learning},
  author={DeepSeek-AI},
  journal={arXiv preprint arXiv:2501.12948},
  year={2025}
}

@Article{Chen2024BGE,
  author={Jianlv Chen and others},
  title={{BGE M3-Embedding}: Multi-Lingual, Multi-Functionality, Multi-Granularity Text Embeddings Through Self-Knowledge Distillation},
  journal={arXiv preprint arXiv:2402.03216},
  year={2024},
  url={https://arxiv.org/abs/2402.03216}
}

@Manual{LangChain2022,
  title={LangChain: Building applications with LLMs through composability},
  author={{LangChain Team}},
  year={2022},
  url={https://github.com/langchain-ai/langchain},
  note={Accessed: 2025}
}
\bibliographystyle{mylatex}
\newpage
\appendix
\onecolumn
\section{Appendix: Detailed Training Configurations}
\label{appendix:training_details}

\subsection{CPT and SFT Stages}
Table~\ref{tab:sft_params} presents the hyperparameters for CPT, SFT-General, and SFT-Combustion. All stages were conducted using the LLaMA-Factory framework with DeepSpeed ZeRO-3.

\begin{table}[h]
\centering
\small
\begin{tabular}{lccc}
\toprule
\textbf{Hyperparameter} & \textbf{CPT} & \textbf{SFT-General} & \textbf{SFT-Combustion} \\
\midrule
Learning Rate & 2.0e-5 & 5.0e-5 & 2.0e-5 \\
LR Scheduler & WSD & Cosine & Cosine \\
Max Length & 16,384 & 16,384 & 20,000 \\
Batch Size (Total) & 256 & 256 & 128 \\ 
Precision & bf16 & bf16 & bf16 \\
Warmup Ratio & 0.01 & 0.05 & 0.03 \\
\bottomrule
\end{tabular}
\caption{Hyperparameters for CPT and SFT Stages.}
\label{tab:sft_params}
\end{table}

\subsection{Reinforcement Learning (GRPO)}
The final stage employed the Group Relative Policy Optimization (GRPO) algorithm using the \texttt{Verl} framework. We utilized vLLM as the rollout engine to efficiently sample multiple responses for reward normalization.

\begin{table}[h]
\centering
\small
\begin{tabular}{lc}
\toprule
\textbf{Parameter} & \textbf{Value} \\
\midrule
Algorithm & GRPO \\
Actor Learning Rate & $2.0 \times 10^{-6}$ \\
Global Training Batch Size & 128 \\
Max Prompt Length & 1,024 \\
Max Response Length & 8,192 \\
KL Coefficient ($\beta$) & 0.005 \\
KL Loss Type & low\_var\_kl \\
Number of Samples ($n$) & 8 \\
Rollout Engine & vLLM \\
Precision & Mixed (bf16/fp32) \\
\bottomrule
\end{tabular}
\caption{Hyperparameters for the GRPO reinforcement learning stage.}
\label{tab:rl_params}
\end{table}

\end{document}